\documentclass[runningheads]{llncs}
\usepackage{graphicx}
\usepackage{cite}
\usepackage{amsmath,amssymb,amsfonts}
\usepackage[ruled,vlined]{algorithm2e}
\usepackage{algorithmic}
\usepackage{efbox,graphicx}
\usepackage{pgfplots}
\pgfplotsset{compat=1.14}
\usepackage{textcomp}
\usepackage{multirow}
\usepackage{soul}
\usepackage{url}

\begin{document}

\title{Improving Self-Organizing Maps with Unsupervised Feature Extraction}
\titlerunning{SOM with Unsupervised Feature Extraction}

\author{Lyes Khacef \and
Laurent Rodriguez \and
Beno\^it Miramond}

\authorrunning{L. Khacef et al.}

\institute{Universit\'e C\^ote d'Azur, CNRS, LEAT, France \\
\email{firstname.lastname@univ-cotedazur.fr}}

\maketitle

\begin{abstract}
The Self-Organizing Map (SOM) is a brain-inspired neural model that is very promising for unsupervised learning, especially in embedded applications. However, it is unable to learn efficient prototypes when dealing with complex datasets. We propose in this work to improve the SOM performance by using extracted features instead of raw data. We conduct a comparative study on the SOM classification accuracy with unsupervised feature extraction using two different approaches: a machine learning approach with Sparse Convolutional Auto-Encoders using gradient-based learning, and a neuroscience approach with Spiking Neural Networks using Spike Timing Dependant Plasticity learning. The SOM is trained on the extracted features, then very few labeled samples are used to label the neurons with their corresponding class. We investigate the impact of the feature maps, the SOM size and the labeled subset size on the classification accuracy using the different feature extraction methods. We improve the SOM classification by $+6.09\%$ and reach state-of-the-art performance on unsupervised image classification.


\end{abstract}


\section{Introduction}
With the fast expansion of Internet of Things (IoT) devices, a huge amount of unlabeled data is gathered everyday. While it is a big opportunity for Artificial Intelligence (AI) and Machine Learning (ML), the difficult task of labeling these data makes Deep Learning (DL) techniques slowly reaching the limits of supervised learning \cite{droniou2015multimodal_perception,chum2019beyond_supervised}. Hence, unsupervised learning is becoming one of the most important and challenging topics in ML.
In this work, we use the Self-Organizing Map (SOM) proposed by Kohonen \cite{kohonen1990som}, an Artificial Neural Network (ANN) that is very popular in the unsupervised learning category \cite{kohonen2001som}. Inspired from the cortical synaptic plasticity and its self-organization properties, the SOM is a powerful vector quantization algorithm which models the probability density function of the data into a set of prototype vectors that are represented by the neurons synaptic weights \cite{rougier2011dsom}.
It has been shown that SOMs perform better in representing overlapping structures compared to classical clustering techniques such as partitive clustering or K-means \cite{budayan2009cluster_vs_som}. 

In addition, SOMs are well suited to hardware implementation based on cellular neuromorphic architectures \cite{sousa2017embedded_som, khacef2018neuromorphic_hardware,rodriguez2018grid_som}. Thanks to a fully distributed architecture with local connectivity amongst hardware neurons, the energy-efficiency of the SOM is highly improved since there is no communication between a centralized controller and a shared memory unit, as it is the case in classical Von-Neumann architectures. Moreover, the connectivity and computational complexities of the SOM become scalable with respect to the number of neurons \cite{rodriguez2018grid_som}. SOMs are used in a large range of applications \cite{kohonen1996som_app} going from high-dimensional data analysis to more recent developments such as identification of social media trends \cite{silva2018social_media}, incremental change detection \cite{nallaperuma2018bahavior_changes} and energy consumption minimization on sensor networks \cite{kromes2019lorawan}.

This work is an extension of the work done in \cite{khacef2019self-organizing_neurons}, where we introduced the problem of post-labeled unsupervised learning: no label is available during training and representations are learned in an unsupervised fashion, then very few labels are available for assigning each representation the class it represents. The latter is called the labeling phase. In \cite{khacef2019self-organizing_neurons}, we used the MNIST dataset \cite{lecun1998mnist} to demonstrate the potential of this unsupervised learning method on the classification problem and compared different training and labelling techniques.
In order to improve the classification accuracy of the SOM and be able to work with more complex datasets, we need to extract useful features from the raw data that will then be classified with the SOM. In the context of unsupervised learning, feature extraction can be done using two different approaches: a classical "machine learning approach" using Sparse Convolutional Auto-Encoders (SCAEs), and a "neuroscience approach" using Spiking Neural Networks (SNNs). The SCAE is trained using gradient back-propagation while the SNN is trained using Spike Timing Dependant Plasticity (STDP). The goal of this work is to compare the performance of both approaches when using a SOM classifier. We also experiment a supervised Convolutional Neural Network (CNN) with the same topology for approximating the best accuracy we can expect from the feature extraction. 

Section \ref{sec_state-of-art} describes the unsupervised feature extraction methods and details the SOM training and labeling algorithms. Then, Section \ref{sec_methods} presents the implementation details of each feature extractor. Next, Section \ref{sec_results} presents the experiments and results on MNIST unsupervised classification. Finally, Section \ref{sec_discussion} and Section \ref{sec_conclusion} discuss and conclude our work.


\section{Related work and methodology}
\label{sec_state-of-art}
In this section, we review the related work and present the proposed methodology. We begin with the unsupervised feature extraction learning part, then how to train the SOM, and we finally explain the labeling procedure.
Our first step is to extract relevant features from the raw data using unsupervised learning.

\subsection{Unsupervised feature extraction}
\subsubsection{Sparse Convolutional Auto-Encoders (SCAEs)}
Introduced by Rumelhart, Hinton and Williams \cite{rumelhart1988autoencoders}, AEs were designed to address the problem of back propagation without supervisor via taking the input data itself as the supervised label \cite{baldi2012autoencoders}. Today, AEs are typically used for dimensionality reduction or weights initialization in CNNs to improve the classification accuracy \cite{masci2011stacked_cae} \cite{kohlbrenner2017pretrain_cnn}. In this work, we want to use AEs as feature extractors with unsupervised learning.
In such cases, the feature map representation of a Convolutional AE (CAE) is most of the time of a much higher dimensionality than the input image. 
While this feature representation seems well-suited in a supervised CNN, the overcomplete representation becomes problematic in an AE since it gives the autoencoder the possibility to simply learn the identity function by having only one weight “on” in the convolutional kernels \cite{masci2011stacked_cae}. 
Without any further constraints, each convolutional layer in the AE could easily learn a simple point filter that copies the input onto a feature map \cite{kohlbrenner2017pretrain_cnn}. While this would later simplify a perfect reconstruction of the input, the CAE does not find any more suitable representation for our data. To prevent this problem, some constraints have to be applied in the CAE to increase the sparsity of the features representation.

The concept of sparsity was introduced in computational neuroscience, as sparse representations resemble the behavior of simple cells in the mammalian primary visual cortex, which is believed to have evolved to discover efficient coding strategies \cite{olshausen1997sparse_coding_v1}. It has been proven that encouraging sparsity when learning the transformed representation can improve the performance of classification tasks \cite{hoyer2004sparseness}. Indeed, the overcomplete architecture of a CAE allows a larger number of hidden units in the code, but this requires that for the given input, most of hidden neurons result in very little activation \cite{ng2011sparse_ae}. 
In a Sparse CAE (SCAE), activations of the encoding layer need to have low values in average. Units in the hidden layers usually do not fire \cite{charte2018tuto_ae} so that the few non-zero elements represent the most salient features \cite{ng2011sparse_ae}.

In order to increase the sparsity of the CAE's feature representation, several methods can be found in the literature. In \cite{masci2011stacked_cae}, the authors use max-pooling to enforce the learning of plausible filters, but the filters are then fine-tuned with supervised learning for the classification. Since we do not want to use any label in the training process, we apply additional constraints in the SCAE, namely weights and activity constraints of types L2 and L1, respectively \cite{jiang2015sparse_penalties}.

\subsubsection{Spiking Neural Networks (SNNs)}
Spiking Neural Networks (SNNs) are a brain-inspired family of ANNs used for large-scale simulations in neuroscience \cite{furber2014spinnaker} and efficient hardware implementations for embedded AI \cite{davies2018loihi}. SNNs are characterized by the spike-based information coding, a computational model of the electrical impulses amongst the biological neurons. The amplitude and duration of all spikes are almost the same, so they are mainly characterized by their emission time \cite{kheradpisheh2018stdp_cnn}. Furthermore, spiking neurons appear to fire a spike only when they have to send an important message, which leads to the fast and extremely energy-efficient neural computation in the brain.

Moreover, SNNs have a great potential for unsupervised learning through STDP \cite{diehl2015stdp}, a biologically plausible local learning mechanism that uses the spike-timing correlation to update the synaptic weights. Kheradpisheh et al. proposed in \cite{kheradpisheh2018stdp_cnn} a SNN architecture that implements convolutional and pooling layers for spike-based unsupervised feature extraction. 
The SNN processes image inputs as follow.
The first layer of the network uses Difference of Gaussians (DoG) filters to detect contrasts in the input image. It encodes the strength of the edges in the latencies of its output spikes, i.e. the higher the contrast, the shorter the latency. 
On the one hand, neurons in convolutional layers detect complex features by integrating input spikes from the previous layer, and emit a spike as soon as they detect their "preferred" visual feature. A Winner-Take-All (WTA) mechanism is implemented so that the neurons that fire earlier perform the STDP learning and prevent the others from firing. Hence, more salient and frequent features tend to be learned by the network. 
On the other hand, neurons in the pooling layers provide translation invariance by using a temporal maximum operation, and help the network to compress the flow of visual data by propagating the first spike received from neighboring neurons in the previous layer which are selective to the same feature.
However, in \cite{kheradpisheh2018stdp_cnn}, the extracted features were classified using a supervised Support Vector Machine (SVM). In this work, we use the unsupervised SOM classifier to keep the unsupervised training from end to end.

\subsection{Unsupervised classification with Self-Organizing Maps (SOMs)}
\subsubsection{SOM learning}
The next step consists in training a SOM using the extracted features. We use a two-dimensional array of $k$ neurons, that are randomly initialized and updated thanks to the following algorithm based on \cite{kohonen1990som}:

\begin{algorithmic}[]
    \STATE
    \STATE \textbf{Initialize} the network as a two-dimensional array of $k$ neurons, where each neuron $n$ with $m$ inputs is defined by a two-dimensional position $p_n$ and a randomly initialized $m$-dimensional weight vector $w_n$.
    \FOR{$t$ from $0$ to $t_f$}
        \FOR{every input vector $v$}
            \FOR{every neuron $n$ in the network}
                \STATE \textbf{Compute} the afferent activity $a_n$ from the distance $d$:
                    \begin{equation}
                    \label{eq_euc-dist}
                        d = \|v - w_n\|
                    \end{equation}
                    \begin{equation}
                    \label{eq_gaussian}
                        a_n = e^{-\frac{d}{\alpha}}
                    \end{equation}
            \ENDFOR
        	\STATE \textbf{Compute} the winner $s$ such that:
            	\begin{equation}
            	\label{eq_max_activity}
                    a_s = \max_{n=0}^{k-1} \left( a_n \right)
                \end{equation}
            \FOR{every neuron $n$ in the network}
            	\STATE \textbf{Compute} the neighborhood function $h_\sigma(t,n,s)$:
                \begin{equation}
                	h_{\sigma}(t,n,s) = e^{-\frac{\|p_n - p_s\|^2}{2{\sigma}(t)^2}}
                \end{equation}
                \STATE \textbf{Update} the weight $w_n$ of the neuron $n$:
                \begin{equation}
                	w_n = w_n + \epsilon(t) \times h_{\sigma}(t,n,s) \times (v - w_n)
                \end{equation}
            \ENDFOR
        \ENDFOR
        \STATE \textbf{Update} the learning rate $\epsilon(t)$:
        \begin{equation}
        	\epsilon(t) = \epsilon_i \left(\frac{\epsilon_f}{\epsilon_i}\right)^{t/t_f}
        \end{equation}
        \STATE \textbf{Update} the width of the neighborhood $\sigma(t)$:
        \begin{equation}
        	\sigma(t) = \sigma_i \left(\frac{\sigma_f}{\sigma_i}\right)^{t/t_f}
        \end{equation}
	\ENDFOR
\end{algorithmic}

It is to note that $t_f$ is the number of epochs, i.e. the number of times the whole training dataset is presented.
The $\alpha$ hyper-parameter is the width of the Gaussian kernel. Its value in Equation \ref{eq_gaussian} is fixed to $1$ in the SOM training, but it does not have any impact in the training phase since it does not change the neuron with the maximum activity. Its value becomes critical though in the labeling process.
The SOM hyper-parameters are reported in Section \ref{sec_results}. 

\subsubsection{SOM labeling}
\label{sec_som-labeling}
The labeling is the step between training and test where we assign each neuron the class it represents in the training dataset. We proposed in \cite{khacef2019self-organizing_neurons} a labeling algorithm based on very few labels. We randomly took a labeled subset of the training dataset, and we tried to minimize its size while keeping the best classification accuracy. Our study showed that we only need $1 \%$ of randomly taken labeled samples from the training dataset for MNIST classification. In this work, we will extend the so-called post-labeled unsupervised learning to SOM classification with features extracted by different means.

The labeling algorithm detailed in \cite{khacef2019self-organizing_neurons} can be summarized in five steps.
First, we calculate the neurons activations based on the labeled input samples from the euclidean distance following Equation \ref{eq_gaussian}, where $v$ is the input vector, $w_n$ and $a_n$ are respectively the weights vector and the activity of the neuron $n$. The parameter $\alpha$ is the width of the Gaussian kernel that becomes a hyper-parameter for the method.
Second, the Best Matching Unit (BMU), i.e. the neuron with the maximum activity is elected.
Third, each neuron accumulates its normalized activation (simple division) with respect to the BMU activity in the corresponding class accumulator, and the three steps are repeated for every sample of the labeling subset.
Fourth, each class accumulator is normalized over the number of samples per class.
Fifth and finally, the label of each neuron is chosen according to the class accumulator that has the maximum activity.
The complete GPU-based source code is available in \url{https://github.com/lyes-khacef/GPU-SOM}.


\section{Implementation details}
\label{sec_methods}
MNIST \cite{lecun1998mnist} is a dataset of $70000$ handwritten digits ($60000$ for training and $10000$ for test) of $28\times28$ pixels.
In order to compare the feature extraction performance, we use the following topologies for the two approaches:  $28\times28\times1 - 64c5 - Xc5 - p5$ for the SCAE and $28\times28\times1 - 64c5 - p2 - Xc5 - p2$ for the SNN, i.e. two convolutional layers of 64 maps and X maps respectively. Each uses $5\times5$ kernels followed by a max-pooling layer. The reason for the different pooling mechanisms is explained in Section \ref{sec_snn-training}. We explore the impact of the number of features X on the classification accuracy.

\subsection{CNN training}
The CNN is modeled in TensorFlow/Keras and trained with Adadelta \cite{zeiler2012adadelta} gradient-based algorithm for 100 epochs with a learning rate of $1.0$. Since the goal is to estimate the maximum accuracy we can expect from each topology, the CNN is trained with the labeled training set by using 10 neurons with a Softmax activation function on top of the last pooling layer. This network is noted as CNN+MLP in the following.

\subsection{SCAE training}
The SCAE is also modeled in TensorFlow/Keras and trained using Adadelta \cite{zeiler2012adadelta} gradient-based algorithm for 100 epochs with a learning rate of 1.0. However, no label is used in the training process, as the goal of the SCAE is to reconstruct the input in the output. The complete SCAE topology is $28\times28\times1 - 64c5 - Xc5 - p5 - u5 - 64d5 - 1d5$, where $u$ stands for up-sampling and $d$ stands for deconvolution (or transposed convolution) layers. The complete architecture is thus symetric. We add to every convolution and deconvoltion layer a weight constraint of type L2, and we add to the second convolution layer that produces the features an activity constraint of type L1. The weights and activity regularisation rates are set to $10^{-4}$. Therefore, the objective function of the SCAE takes in account both the image reconstruction and the sparsity constraints.

\subsection{SNN training}
\label{sec_snn-training}
The SNN is modeled in SpykeTorch \cite{mozafari2019spyketorch}, an open-source simulator of convolutional SNNs based on PyTorch \cite{paszke2019pytorch}. The SNN is trained with STDP layer by layer, with a different pooling mechanism than the CNN and SCAE. Except for the number of feature maps and kernel sizes, we kept the same hyper-parameters as the original implementation of \cite{kheradpisheh2018stdp_cnn} that can be found on \cite{mozafari2019spyketorch}. Hence, we used a pooling layer of $2\times2$ after each convolutional layer, with a padding of $1$ before the second convolutional layer. The threshold of the neurons in the last convolutional layer were set to be infinite so that their final potentials can be measured \cite{kheradpisheh2018stdp_cnn}. 
Finally, the global pooling neurons compute the maximum potential at their corresponding receptive field and produce the features that will be used as input for the SOM. 
Our experimental study showed that the added padding and the pooling mechanism proposed in \cite{mozafari2019spyketorch} performs better than the one used in the CNN and SCAE (i.e. no pooling and one polling layer), with a gain of $1.43\%$ on the maximum achievable accuracy.


\section{Experiments and results}
\label{sec_results}
The SOM training hyper-parameters were found with a grid search: $\epsilon_i = 1.0$, $\epsilon_f = 0.01$, $\eta_i = 10.0$, $\eta_f = 0.01$, $\alpha = 1.0$ and the number of epochs is $10$.

\begin{figure}[h!]
	\centerline{\efbox{\includegraphics[width=0.7\linewidth]{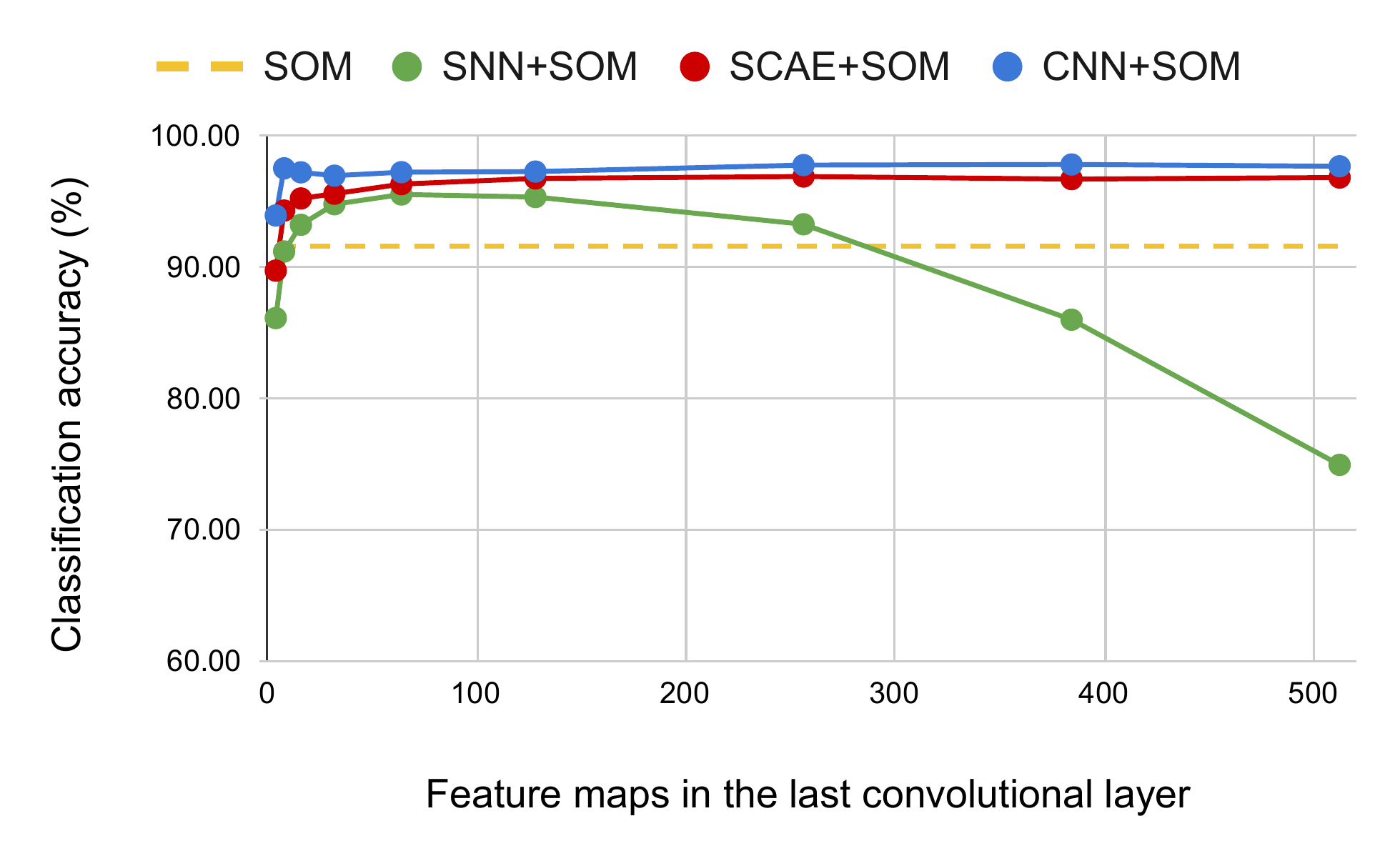}}}
	\caption{SOM classification accuracy using CNN, SCAE and SNN feature extraction vs. number of feature maps with 256 SOM neurons and 10\% of labels.}
	\label{fig_deepsom-maps}
\end{figure}

First, figure \ref{fig_deepsom-maps} shows the impact of the number of feature maps in the second convolutional layer, using 256 neurons in the SOM and 10\% of labels. We deliberately use a large number of labels to avoid any bias due to the labeling performance, and focus on the impact of the feature maps. The accuracy of the CNN+SOM and SCAE+SOM is increasing with respect to the number of feature maps, reaching a maximum at 256 maps.
Interestingly, the CNN+SOM performs better with 8 maps (97.56\%) than with 16 (97.25\%), 32 (97.00\%), 64 (97.26\%) or 128 (97.31\%) maps. This is due to the tradeoff between additional information and additional noise induced by more feature maps according to the SOM classification. In fact, the CNN+MLP supervised baseline accuracy is increasing from 98.7\% to 99\% when the feature maps increase from 8 to 512. 
This observation is more pronounced when we look at the SNN+SOM that reaches a maximum accuracy for 64 maps then drastically decreases with more feature maps. Following the approach of \cite{kheradpisheh2018stdp_cnn}, we used a SNN+SVM supervised baseline and its accuracy increases from 97\% to 98\% when the feature maps increase from 64 to 512. It means that the increasing number of feature maps for the SNN produces noisy features that do not affect the supervised classification but do decrease the unsupervised classification accuracy, because the SOM prototypes overlap and become less descriminative. 
Thus, we choose 256 maps for the CNN and SCAE that produce a feature size of 4096, and 64 maps for the SNN that produces feature maps of size 3136. We remark that the SNN features size is different from the CNN/SCAE features size, which is due to the to the added padding and the different pooling mechanism as explained in Section \ref{sec_snn-training}.

\begin{figure}[h!]
	\centerline{\efbox{\includegraphics[width=0.7\linewidth]{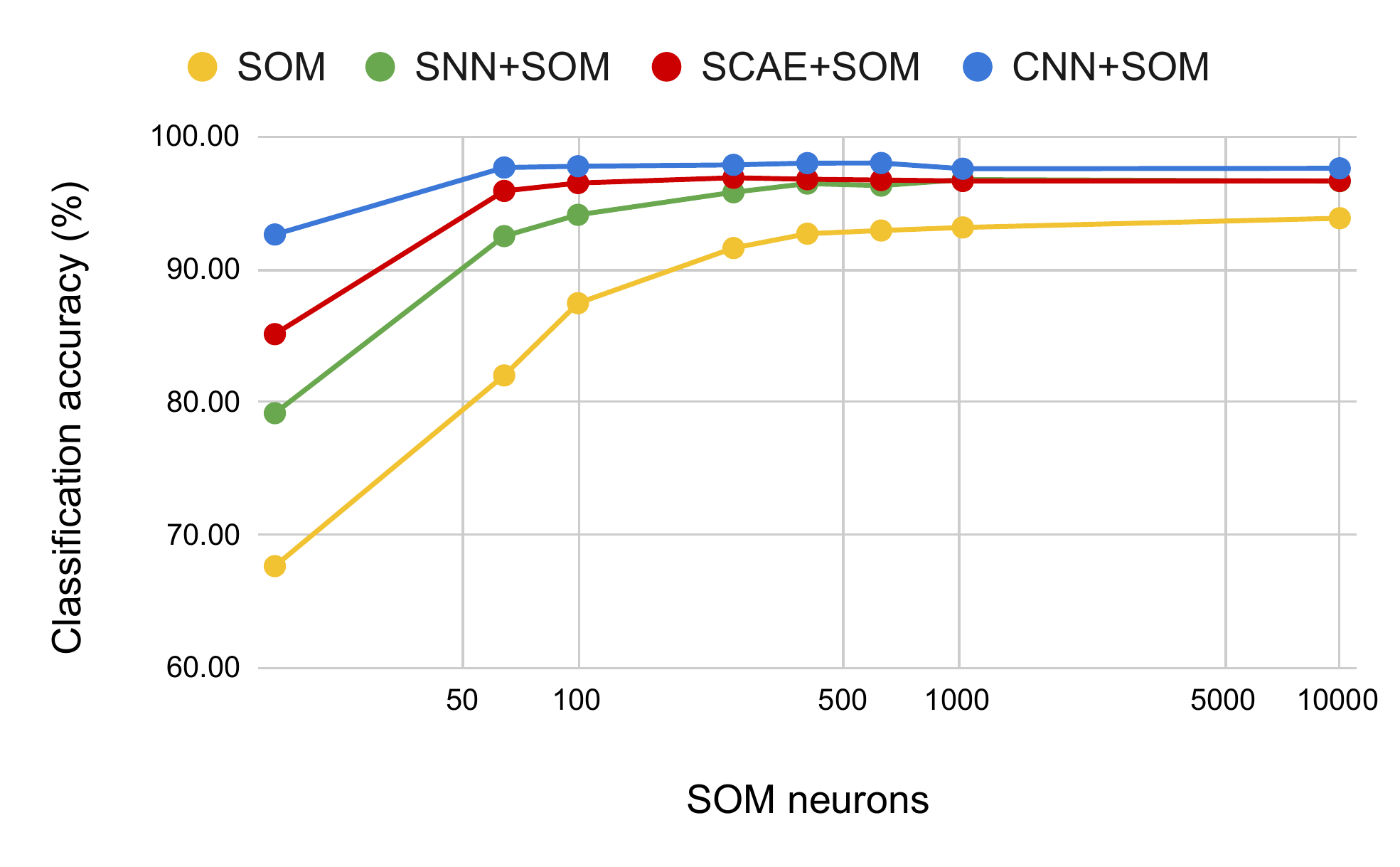}}}
	\caption{SOM classification accuracy using CNN, SCAE and SNN feature extraction vs. number of SOM neurons with the optimal topologies and 10\% of labels.}
	\label{fig_deepsom-neurons}
\end{figure}

Second, with the above mentioned topologies, we investigate the impact of the SOM size with 10\% of labels, from 16 to 10000 neurons. We see in Figure \ref{fig_deepsom-neurons} that the accuracy of the four systems is increasing with respect to the number of neurons. We notice that the SNN-SOM reaches the same accuracy as the SCAE+SOM starting from 1024 neurons. Nevertheless, for the next step of the study, it is important to keep the same number of neurons. Hence, we have chosen the number of neurons for which one of the SCAE+SOM or SNN+SOM reaches the maximum accuracy, which is equal to 256 neurons with respect to the SCAE+SOM accuracy.

\begin{figure}[h!]
	\centerline{\efbox{\includegraphics[width=0.7\linewidth]{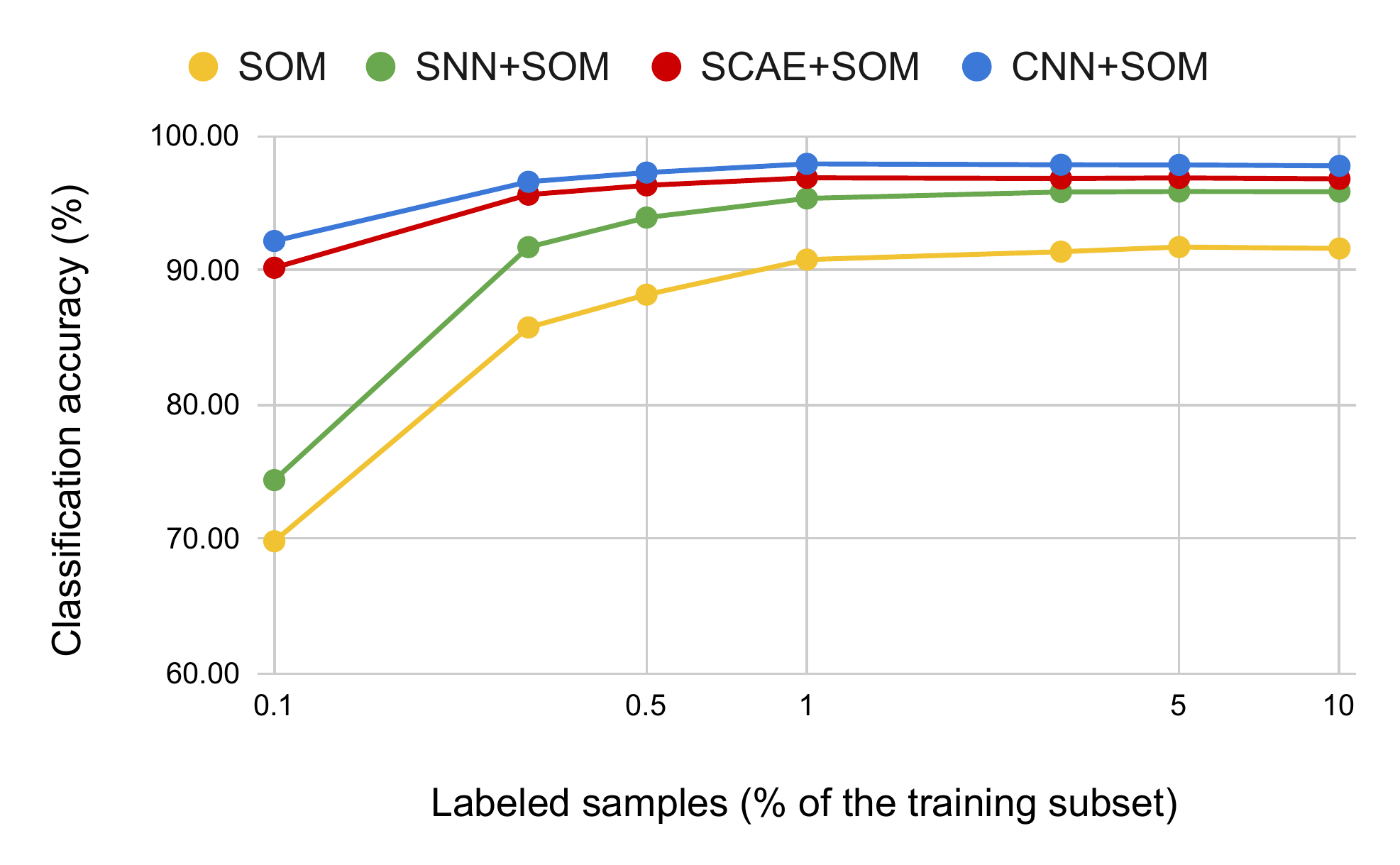}}}
	\caption{SOM classification accuracy using CNN, SCAE and SNN feature extraction vs. \% of labeled data from the training subset for the neurons labeling with the optimal topologies and 256 SOM neurons.}
	\label{fig_deepsom-labels}
\end{figure}

Third, using 256 neurons for the SOM, we investigate the impact of the labeling subset size in terms of \% of the training set. Figure \ref{fig_deepsom-labels} shows that the accuracy increases when the labeled subset increases. Interestingly, the CNN+SOM and SCAE+SOM reach their maximum accuracy with only 1\% of labeled data, while the SNN+SOM and SOM need approximately 5\% of labeled data. Since the SCAE+SOM performs better than the SNN+SOM, we only need 1\% of labeled data. It confirms the results obtained in \cite{khacef2019self-organizing_neurons}.

\begin{figure}[h!]
	\centerline{\efbox{\includegraphics[width=0.7\linewidth]{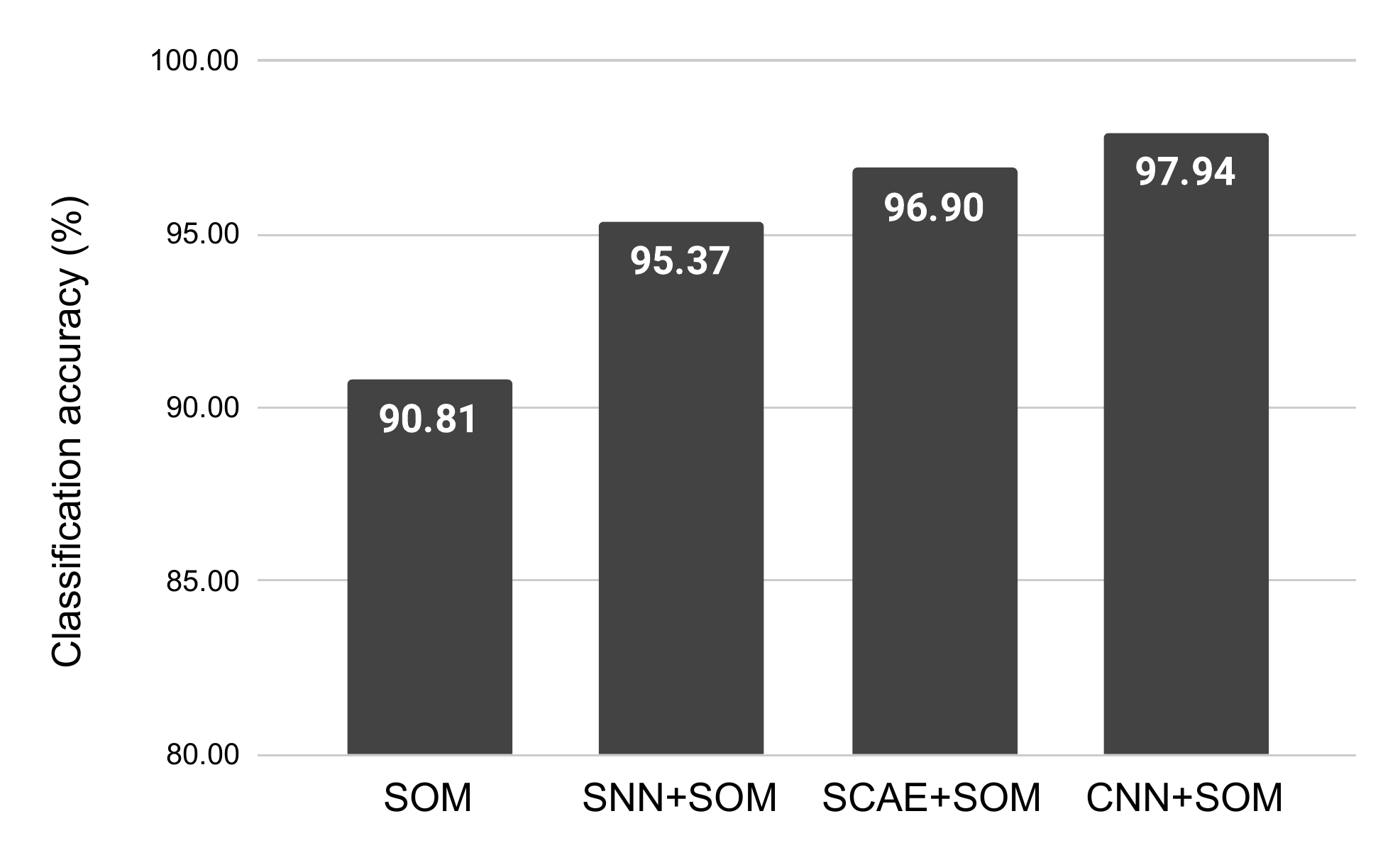}}}
	\caption{SOM classification accuracy using CNN, SCAE and SNN feature extraction vs. summary of the comparative study with the optimal topologies, 256 SOM neurons and 1\% of labels.}
	\label{fig_deepsom-accuracy}
\end{figure}

Finally, the comparative study of the four settings with the best topology of each, using 256 neurons for the SOM and 1\% of labeled data for the neurons labeling is summarized in Figure \ref{fig_deepsom-accuracy}. As expected, the SOM without feature extraction has the worst accuracy of $90.91\% \pm 0.15$ and the CNN+SOM with supervised feature extraction reaches the best accuracy of $97.94\% \pm 0.22$. More interestingly, with fully unsupervised learning, the SCAE performs better than the SNN ($+1.53\%$), with $96.9\% \pm 0.24$ and $95.37\% \pm 0.58$ respectively.


\section{Discussion}
\label{sec_discussion}

\begin{table}[ht]
\centering
\caption{Comparison of unsupervised feature extraction and classification techniques in terms of accuracy and hardware cost.}
\label{tab_cost}
\begin{center}
\resizebox{0.95\linewidth}{!}{
    \begin{tabular}{c c c c c c c}
    \hline
    \multicolumn{2}{c}{\textbf{Feature extraction}} & \multicolumn{2}{c}{\textbf{Classification}} & \multicolumn{3}{c}{\textbf{Performance}}                             \\
    \textbf{Model}         & \textbf{Learning}        & \textbf{Model}      & \textbf{Learning}      & \textbf{Accuracy (\%)} & \textbf{Error (\%)} & \textbf{Hardware cost} \\ \hline
    CNN                    & Supervised               & MLP                 & Supervised             & 99.00                  & 1.00                & High                   \\
    CNN                    & Supervised               & SOM                 & Unsupervised           & 97.94                  & 2.06                & Medium                 \\
    SCAE                   & Unsupervised             & SOM                 & Unsupervised           & 96.90                  & 3.10                & Medium                 \\ 
    SNN                    & Unsupervised             & SOM                 & Unsupervised           & 95.37                  & 4.63                & Low                    \\ \hline
    \end{tabular}
}
\end{center}
\end{table}

Table \ref{tab_cost} shows the gap between supervised and unsupervised methods for feature extraction and classification. Interestingly, we only lose about $1\%$ of accuracy when going from CNN+MLP to CNN+SOM, and another $1\%$ when going from CNN+SOM to SCAE+SOM. The gap is slightly higher when going from SCAE+SOM to SNN+SOM, which is about $1.5\%$. In return, the hardware cost decreases when using SOMs and SNNs, thanks to the brain-inspired computing paradigm (distributed and local).
Indeed, we showed in \cite{khacef2018mlp_vs_snn} that the SNN has a gain of approximately $50\%$ in hardware resources and power consumption when implemented in dedicated FPGA and ASIC hardware.

\begin{table}[h]
\centering
\caption{MNIST unsupervised learning with AE-based feature extraction: state of the art reported from \cite{ji2018invariant_clustering}.}
\label{tab_mnist}
\begin{center}
\resizebox{0.6\linewidth}{!}{
    \begin{tabular}{l c}
    \hline
    \textbf{Method}                      & \textbf{Accuracy (\%)}  \\ \hline
    AE + K-means \cite{bengio2006greedy_training}                          & 81.2                    \\
    Sparse AE + K-means \cite{ng2011sparse_ae}                  & 82.7                   \\
    Denoising AE + K-means \cite{vincent2010stacked_denoise_ae}               & 83.2                    \\
    Variational Bayes AE + K-means \cite{kingma2013variational_ae}       & 83.2                    \\
    SWWAE + K-means \cite{zhao2015stacke_ww_ae}                      & 82.5                    \\
    Adversarial AE \cite{makhzani2015adversarial_ae} & 95.9                   \\
    Sparse CAE + SOM [Our work]       & \textbf{96.9}          \\ \hline
    \end{tabular}
}
\end{center}
\end{table}

Overall, the SCAE+SOM reaches the best accuracy of $96.9\% \pm 0.24$ on MNIST classification with unsupervised learning. As shown in Table \ref{tab_mnist}, we achieved state of the art accuracy compared to similar works that followed an AE-based approach.
The sparsity constraints of the SCAE through the weights and activities regularization significantly improved the SOM classification accuracy. Indeed, without these constraints, the CAE+SOM with the same configuration achieves an accuracy of $94.9\% \pm 0.24$, which means a loss of $-2\%$.

A similar study was conducted in \cite{falez2019feature_learning}, but it was limited to one layer SCAE and SNN, and a supervised SVM was used for classification. The authors concluded that the SCAE reaches a better classification accuracy. Our study extands their finding to multiple convolutional layers by using unsupervised learning for both feature extraction and classification.
Nevertheless, the SNN+SOM remains attractive due to the hardware-efficient computation of spiking neurons \cite{khacef2018mlp_vs_snn} associated to the cellular neuromorphic architecture of the SOM \cite{rodriguez2018grid_som}.


\section{Conclusion and further works}
\label{sec_conclusion}
In the context of unsupervised learning, we conducted a comparative study for unsupervised feature extraction, and concluded that the SCAE+SOM achieves a better accuracy thanks to the sparsity constraints that were applied to the SCAE through weights and activities regularization. However, the SNN+SOM remains interesting due to the hardware efficiency of spiking neurons. We achieved state of the art performance on MNIST unsupervised classification, using post-labeled unsupervised learning with the SOM. The future works will focus on using the feature extraction on more complex datasets to improve the accuracy of a multimodal unsupervised learning mechanism \cite{khacef2020cdz} based on SOMs.


\section*{Acknowledgment}
This material is based upon work supported by the French National Research Agency (ANR) and the Swiss National Science Foundation (SNSF) through SOMA project ANR-17-CE24-0036.


\end{document}